\documentclass[letterpaper]{article} 
\usepackage{aaai24}  
\usepackage{times, enumitem}  
\usepackage{helvet}  
\usepackage{courier}  
\usepackage[hyphens]{url}  
\usepackage{graphicx} 
\usepackage{natbib}  
\usepackage{caption} 
\usepackage{algorithm}
\usepackage{algpseudocode}
\usepackage{caption}
\usepackage{amsmath}
\usepackage{subcaption}

\usepackage{newfloat}
\usepackage{listings}
\DeclareCaptionStyle{ruled}{labelfont=normalfont,labelsep=colon,strut=off} 
\floatstyle{ruled}
\newfloat{listing}{tb}{lst}{}
\floatname{listing}{Listing}
\title{AI-Based Teat Shape and Skin Condition Prediction for Dairy Management}
\author{
    Yuexing Hao \equalcontrib \thanks{corresponding author},
    Tiancheng Yuan \equalcontrib,
    Yuting Yang \equalcontrib,\\
    Aarushi Gupta,
    Matthias Wieland,
    Ken Birman,
    Parminder S. Basran
}
\affiliations{

    Ithaca NY, USA\\
    \{yh727, ty373, yy354, ag2288, mjw248, psb92, kpb3\}@cornell.edu
}

\usepackage{bibentry}

\begin{document}
\maketitle

\begin{abstract}
Dairy owners spend significant effort to keep their animals healthy. There is good reason to hope that technologies such as computer vision and artificial intelligence (AI) could reduce these costs, yet obstacles arise when adapting advanced tools to farming environments. In this work, we adapt AI tools to dairy cow teat localization, teat shape, and teat skin condition classifications. We also curate a data collection and analysis methodology for a Machine Learning (ML) pipeline. The resulting teat shape prediction model achieves a mean Average Precision (mAP) of 0.783, and the teat skin condition model achieves a mean average precision of 0.828. Our work leverages existing ML vision models to facilitate the individualized identification of teat health and skin conditions, applying AI to the dairy management industry. 



\end{abstract}
\section*{Keywords}
Digital Agriculture, Machine Learning, Computer Vision, Decision Support Under Uncertainty, Applications

\section{Introduction}

Traditionally, dairy cow teat health assessment requires close examination by a trained professional. Although veterinarians routinely perform this task as part of dairy clinical practice, dairy workers in small farms find the task time-consuming, reducing the accessibility of a valuable predictive tool. On large farms, individualized teat health assessments are impractical: thousands of cows might be managed by a few dozen workers. Yet daily examination of cow teat health could help identify changes that might be early precursors of animal health issues. Our work focuses on dairy cow teat health assessment through the 1) automated and accurate teat shape assessment, and 2) creation and deployment of computer vision. 
 
There has been limited research on machine-learning techniques for solving this problem even in rotary milking parlors with excellent lighting, good animal separation, and high-quality animal identification.  One widely cited effort studied cow teat condition classification from a veterinary perspective ~\cite{evaluationBovine}, but focused on clinical settings and did not consider the use of machine learning models for identifying teat shape. Our project provides a more comprehensive machine learning solution for use in milking parlors.  Here we report on data collection, preparation of training data sets labeled with domain-expert knowledge, development of fully-trained ML models, and assessment of its performance using data from commercial farms. 

A well-known concern about ML is that training models can be prohibitively expensive.   Unusually, our approach avoids the need to undertake model training from the ground up.  We evaluated a variety of preexisting open-source computer-vision models, identifying one model that had good baseline performance.  We then performed fine-tuning of its model parameters and conducted additional training with our labeled data, obtaining a refined open-source model that can perform cow teat localization, teat shape, and skin condition classification with high accuracy and yet at low cost.

Accordingly, this paper focuses on three questions:
\begin{enumerate}
    \item Can we obtain high quality still images ({\em keyframes}) from fixed video cameras in a rotary milking parlor? 
    \item Given a choice of images for one cow, can we select the image that best visualizes the stall-id and the cow's teats?
    \item Can we accurately classify teat shape and skin condition?
\end{enumerate} 

Answering these questions will contribute to dairy science in several ways.  In a practical sense, our work is a step towards routine monitoring of teat shape and teat skin condition in a medium-size dairy farm, enabling us to study the actual value of this sort of information. We hypothesize that deploying our ML models could improve dairy herd management, pinpoint issues that arise, and enable timely intervention to head off mastitis or prevent the spread of potentially contagious pathogens, but follow-up studies of deployed solutions will be needed to validate or refute this belief. Our approach additionally yields data suitable for inclusion into repositories that could be used to develop follow-on machine-intelligent solutions (such as for evaluating animal gait and to sense evidence of discomfort), even as we also use to further refine our models.  We also hope to extract a variety of metrics for dairy productivity, which would be valuable when optimizing farm performance.

Production deployment of our ML solution still lies in the future: this paper is  focused on the ML tools themselves.  As noted, our approach can be carried out on standard laptops because we based our solutions on existing open-source, off-the-shelf AI vision tools.  This contrasts with past approaches that required data-center scale computing resources and were environmentally problematic.  Moreover, we expose trade-offs that other researchers with similar goals might encounter by identifying and resolving the practical problems that arise when deploying ML solutions into a rotary milking parlor.  As an example, we find that there are only a few locations at which cameras can conveniently be placed, and identify timing constraints that would arise if an immediate response to a teat condition (such as spraying a medicinal solution) should occur before the animal leaves the parlor.  For each identified question, we  discuss our proposed solutions, lowering the bar to further work in this domain. 


\section{Scientific Background}

\subsection{Dairy Cow Teat Health Metrics}
Dairy cow teat condition is widely used as a predictor of animal health and anticipated milk quality ~\cite{morphology, WIELAND201811447, seykora1985udder}. 
Poor or gradually degrading teat health is recognized as a risk of {\em mastitis:} one of the most important dairy diseases due to its harmful consequences for farm productivity \cite{ruegg2003investigation}. Mastitis prevention strategies typically focus on two approaches: minimizing bacterial presence at the teat end and enhancing the cow's natural resistance to these pathogens \cite{hogeveen2011current}. Studies have shown that teat-end shape is correlated with a cow's resistance to developing mastitis  ~\cite{teatShapOrifice}, somatic cell count and milkability ~\cite{seykora_heritabilities_1985, wieland2017longitudinal}. 

To create a ground-truth data set for teat condition classification, our team works with veterinarians and veterinary assistants, who supervise certain milking sessions, manually scoring each cow's teats with respect to shape and skin condition. The scoring metrics used for teat shape assessment are based on Seykora and Daniel ~\cite{morphology} guidelines, wherein teat shape is scored as [1: pointed, 3: flat, 7: round-flat, 8: round-ring]. For skin condition assessment, the veterinary team employed Neijenhuis \cite{evaluationBovine} guidelines, scored as:  [1: normal skin, 3: teat with open lesion].  

In a clinical setting, visual teat analysis would be supplemented by tactile assessments.  There are other condition scoring dimensions that could be performed, including evidence of hyperkeratosis \cite{hillerton2005teat}, presence of hock lesions \cite{kielland2009prevalence}, quality of lower leg hygiene, quality of udder hygiene \cite{schreiner2003relationship, cook2007tool}, and presence of skin-open lesions.  All of these are important in clinical mastitis risk health assessment, and our future work will
need to explore, although physical manipulation of the teats would not be practical in our setting, hence we would need to explore other traits that track the evolution of teat condition over time, such as redness/swelling and painful reaction to contact with the milking equipment.

\subsection{Machine Learning for Dairy Health Management}
Our effort contributes within the broader area of technology development for dairy farm automation and management.  The area is active, and includes prior work that studied, evaluated, and deployed machine learning techniques for tasks that include overall farm management (nutrition, hydration, animal activity), herd reproduction management, and animal behavior analysis \cite{ml_in_dairy, ml_in_dairy_review, Disease_Detection}.  Many in the field are arguing that the future dairy farm could be reconceived as having a cyber counterpart (sometimes called a {\em digital twin}), in which the farm is modelled as a generator of many distinct data streams, each with its own purpose and data formats, and each used to train and then trigger a specialized task-specific model or database functionality \cite{gupta_digital_2024}.


\begin{figure}[htb]
\centering
\includegraphics[width=0.9\linewidth]{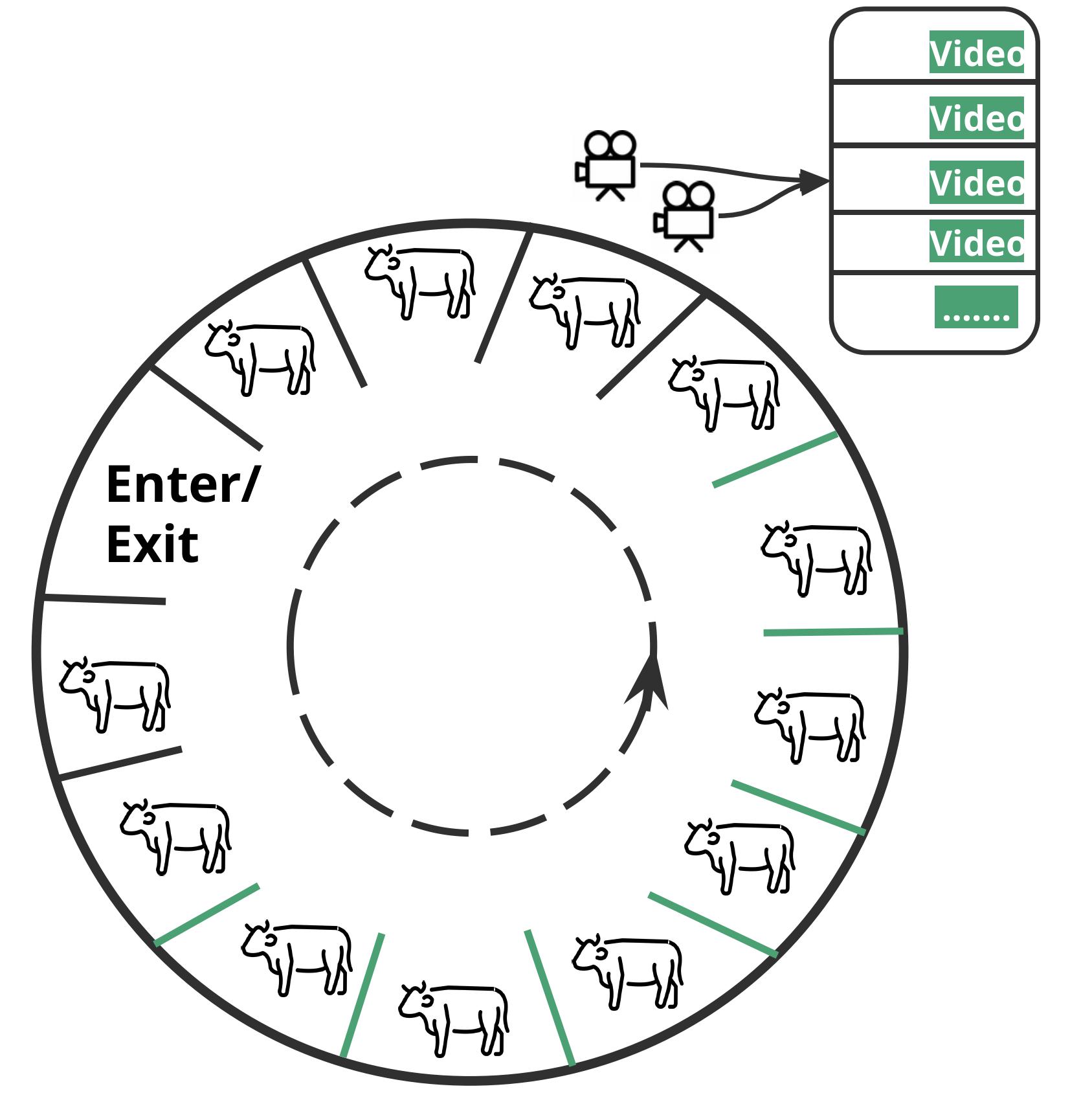}
\caption{Milking parlor and duo-camera setting illustration}
\label{fig:parlor_duo_cam}
\end{figure}

Dairy cows must be identified when entering the rotary milking parlor so the milking data can be obtained from each cow and integrated with the existing dairy information management system.  Currently, this is done using numbered ear tags, Radio Frequency Identification (RFID), and (as needed) human visual inspection.  Our work does not currently
explore options for augmenting these with computer vision tools, but such a step is certainly a possibility for future investigation.


Given an identified animal, two data types can be used as inputs to a machine-learning pipeline.  One category consists of numerical (tabular) data.  Numerical metrics can be captured using sensors, laboratory reports, and milk quantity measurements.  The resulting data set can then be 
used to train models for assessing health metrics, such as heat stress \cite{heat_stress_ml}, estrus \cite{ml_estrus}, mastitis \cite{mastitis_ml} prediction, and behavioral analysis \cite{behavioral_ml} to assist dairy management. For example, \cite{ml_estrus} utilized cow's activity and temperature data in their LCE algorithm that enables automatic estrus detection.  \cite{mastitis_ml} integrated data from cow's health records to develop machine learning models for early prediction of clinical mastitis. The real-time system utilizes these models through data integration across sensors and other data sources to provide analysis and information that help decision-making \cite{numerical_dairy_system}.  

The second pipeline involves images and other image-like data such as ultrasound. For example, computer vision models have been developed that can produce a Body Condition Scoring (BCS) metric, computed by analysis of two-dimensional or three-dimensional photos or videos, thermal images, and even by fusing multiple imaging modalities by capturing simultaneous information using more than one imaging device \cite{bcs_dev, bcs_3d, bcs_body_shape}. Vision-based machine learning models can be trained for tasks such as identifying individual dairy cows, categorizing feeding behavior monitoring \cite{visual_identification_feeding_behavior}, and labeling body parts in a full animal image \cite{cow_body_detection}. 
In future work we will link the two kinds of data to arrive at a single holistic perspective on animal health that integrates all forms of information and tracks temporal evolution of animal health. 

\begin{figure} [H]
\centering
\includegraphics[width=9cm]{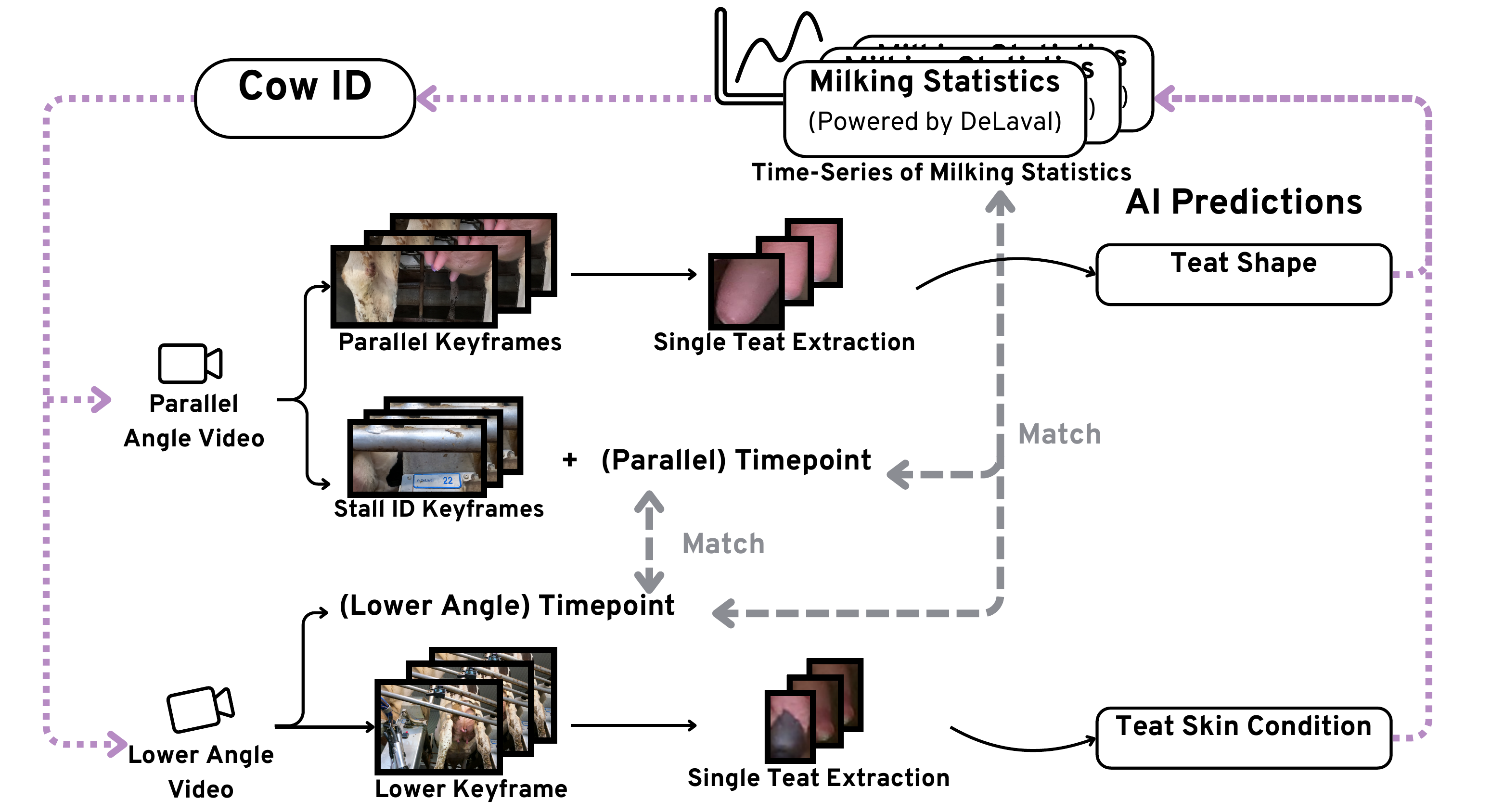} 
\caption{Overall System Workflow}
\label{fig:SystemOverview}
\end{figure}

As noted earlier, most prior research on the use of ML in dairy animal health assessment has occurred in clinical settings, where a veterinarian is examining a single animal~\cite{teat_udder_ml, porter_feasibility_2021, gupta2024digital}. In Slob et al.'s systematic review of ML applications on dairy farm management, teat health classification is the most heavily used ML metric, and mastitis detection is the most important task dependent upon the assessment results~\cite{slob2021application}. Also relevant are clinical tools that can assess teat conditions for individual animals \cite{BASRAN202010703}.  Although our work explores ideas motivated by these clinical tools, we believe that the long term future will tend to differentiate routine health management of the herd (``outside the clinic'') from the types of tools and tests performed in clinical environments.



\section{Data Processing}

\subsection{Data Collection}

We collected video datasets from an Upstate New York dairy farm on October 9th, 2023. The video streams were captured using dual GoPro cameras positioned at lower and parallel angles relative to the cow teats. The veterinarian (a milk quality and udder health specialist with 17 years of experience as bovine veterinarian, certifications: Dip. ECBHM, PhD, DVM) scored the teat shape and skin condition manually, following the Seykora and Daniel \cite{seykora_heritabilities_1985, evaluationBovine} guidelines. 

Although a GoPro captures video, the video data stream itself consists of a series of still images called {\em keyframes} separated by zero or more {\em delta} frames.  For our work, we limited consideration to the key frames.  We disable GoPro data compression and automated image touchup: any image transformation could conceal a teat condition issue much as makeup and digital transformations can conceal skin defects or artificially manipulate an actor's appearance in a movie. 

As shown in Figure~\ref{fig:parlor_duo_cam}, the milking parlor consists of a series of stalls that move slowly in a circle.  The cow enters for premilking teat preparation, is milked,  then released back into the dairy herd. Our cameras are fixed in place and continuously record video of the cows' teats and udders as the parlor rotates past.  This yields multiple images of each animal after milking, but while still in the rotary parlor (Green stalls, Figure \ref{fig:SystemOverview}).  Our camera position allowed for an automated response to the analysis, provided the assessment is completed within one to two seconds.

\subsection{Data Labeling}

Traditionally, computer vision training starts with the acquisition and annotation of comprehensive image datasets that often have hundreds of thousands of examples.  In contrast, our work adopts a preexisting computer vision model trained on very general data but then additionally trains it for the daily task.  Thus, our focus is on aspects specific to dairy teat health assessment.  We start by selecting high-quality keyframe images from the data set collected from the farm.   This selection process discards images where teats are difficult to distinguish, with blurring or poor lighting and motion effects.  For training purposes, our veterinary experts considered only the selected data, annotating a portion that we used to refine the vision model's ability to detect the teats, classify teat end shape, and assess teat skin condition.   

Data preparation is carried out using a package called LabelMe\footnotemark.  
LabelMe output takes the form of JSON files containing annotation details for each image in a dataset \cite{russell2008labelme}. To conform to the standard COCO (Common Objects in Context) object detection dataset format \cite{lin2014microsoft}, a format favored in many deep learning frameworks, we then implement a custom aggregation process that consolidates these annotation files into cohesive datasets.
\footnotetext{http://labelme.csail.mit.edu/Release3.0/}
Data consolidation involves the development of a tailored script to systematically collate annotation data from the individual JSON files generated by LabelMe. The resulting dataset is organized into two comprehensive JSON files: one intended for use during model fine-tuning (training), and the other for validation. A conventional train-test split is applied, with 90\% of the data allocated for model training and the remaining 10\% used for validation.


\begin{figure*}
    \centering
     \begin{subfigure}[b]{0.4\textwidth}
     \centering
     \includegraphics[width=1\textwidth]{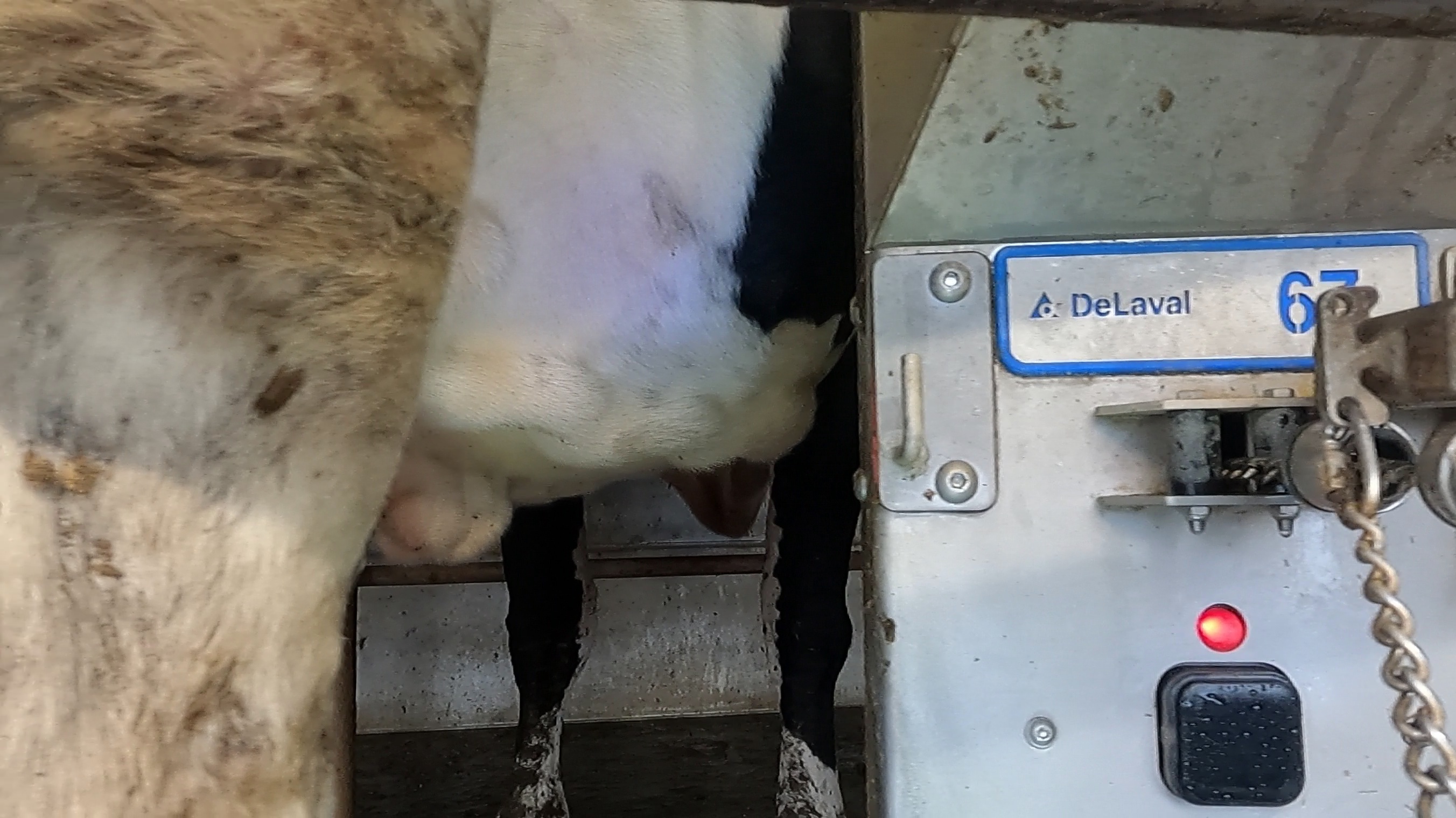}
     \caption{Example of non-keyframe}
     \label{fig:wrong_stallid}
     \end{subfigure}
     \begin{subfigure}[b]{0.4\textwidth}
     \centering
     \includegraphics[width=1\textwidth]{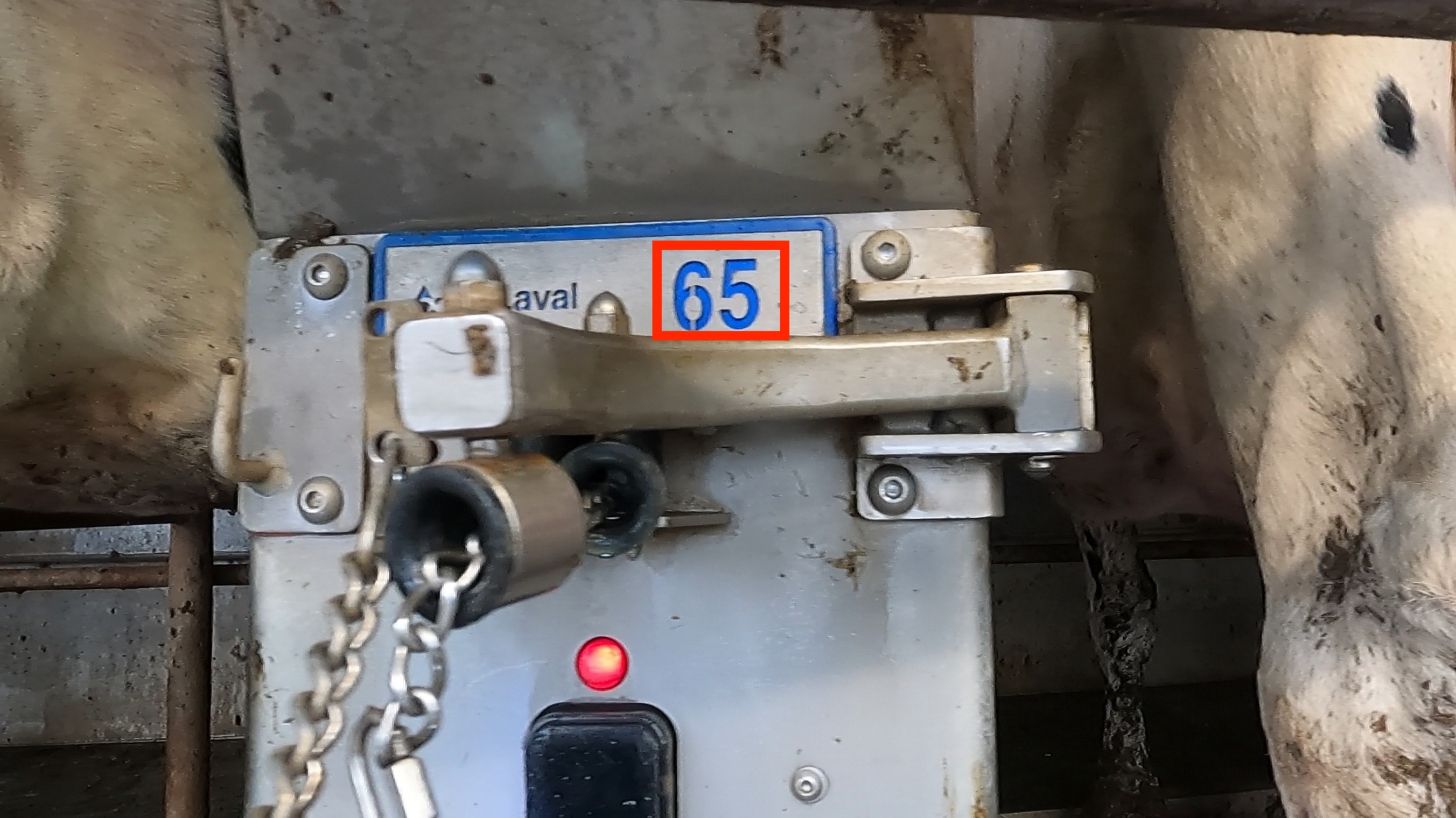}
     \caption{Example keyframe of stall ID}
     \label{fig:kf_stallid}
     \end{subfigure}
     \begin{subfigure}[b]{0.4\textwidth}
     \centering
     \includegraphics[width=1\textwidth]{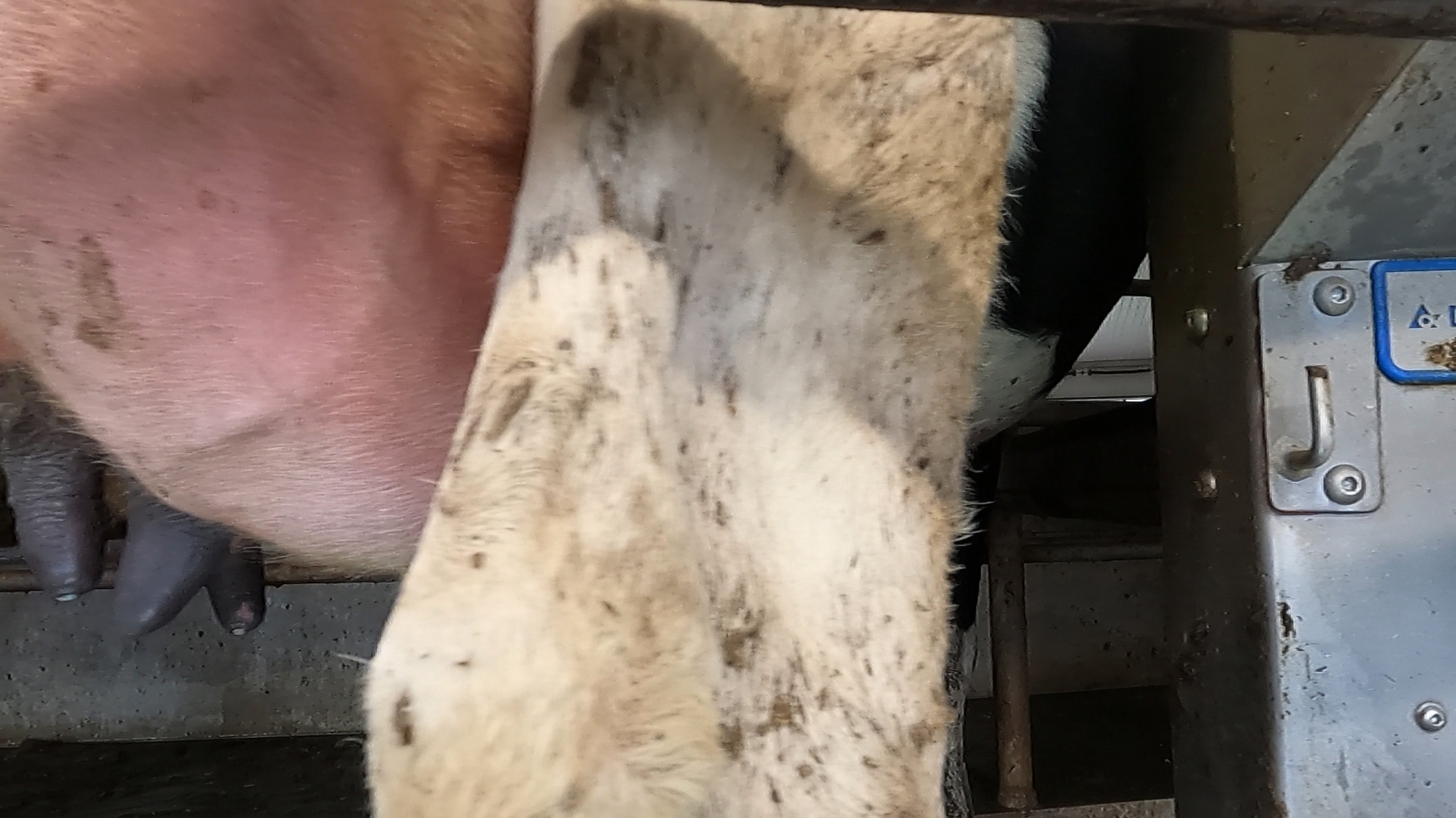}
     \caption{Example non-keyframe of cow Teat}
     \label{fig:wrong_cowteat}
     \end{subfigure}
     \begin{subfigure}[b]{0.4\textwidth}
     \centering
     \includegraphics[width=1\textwidth]{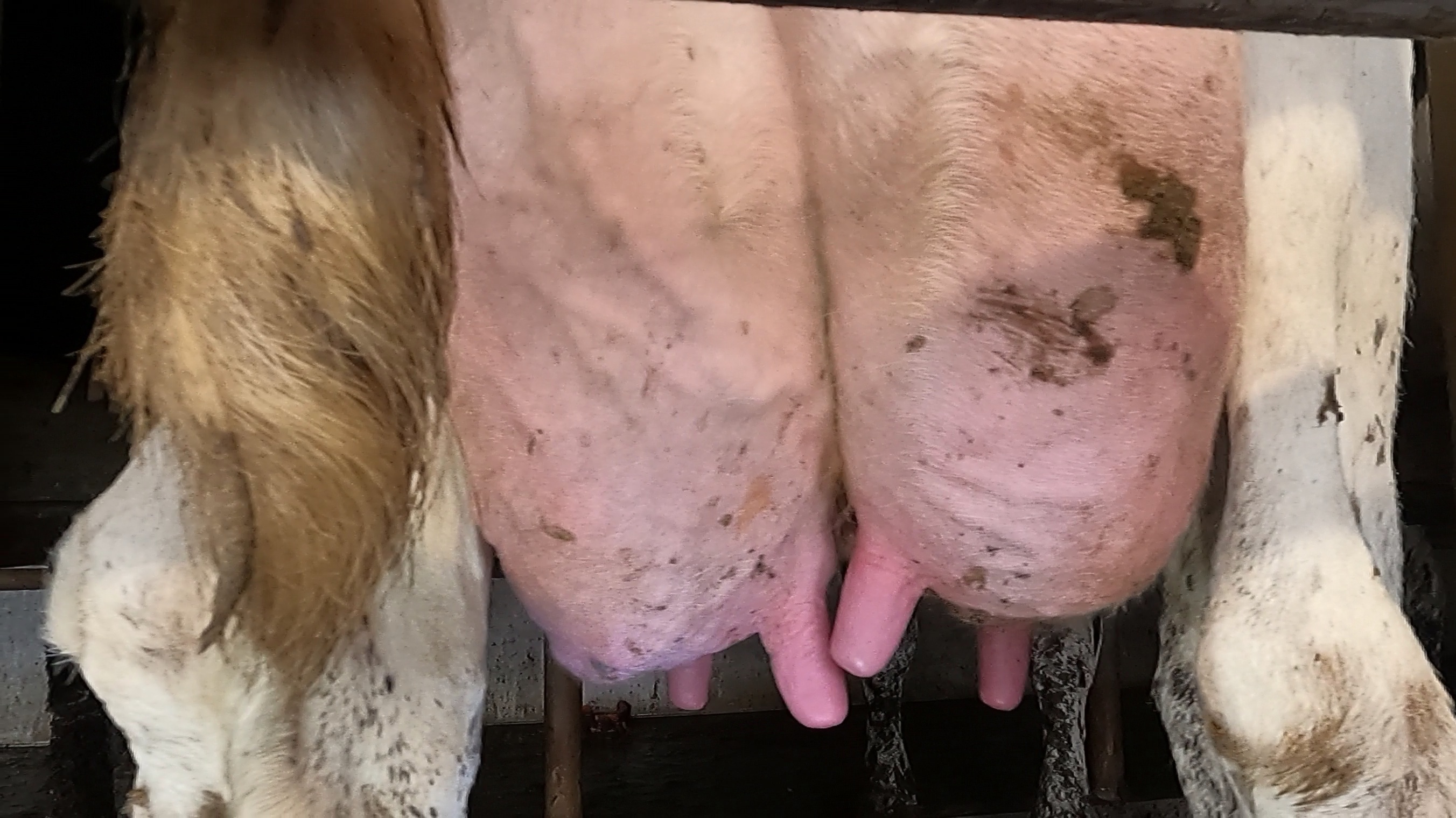}
     \caption{Example keyframe of cow Teat}
     \label{fig:kf_cowteat}
     \end{subfigure}

\caption{Examples of keyframes to be processed}
\label{fig:keyframes}
\end{figure*}

\subsection{Automated Keyframe Selection}
\label{sec:key_frame_extraction}

The first step is to create an ML specialized in evaluating image quality within a stream of keyframes.  There are two subtasks: (1) identification of images that include the cow's stall ID; (2) selection of 2-3 high-quality teat images. These both occur on the same video segment, which shows an individual cow for approximately 3 seconds each.  

Data selection proves to be surprisingly challenging. 
As an example, consider the identification of the stall ID.  Even if an image contains an ID tag, it could be out of focus, the tag may be obstructed, or the frame may capture half of it as the parlor rotates. For example, Figure~\ref{fig:wrong_stallid} is a frame in which the stall ID tag is blocked by the milking parlor device. Accordingly, the algorithm uses two criteria for the frame selection (1) high confidence from the Optical Character Recognition (OCR) model; (2) if the location of the tag is not on the left or right edge in the frame, which is likely to truncate out part of the number. Figure~\ref{fig:kf_stallid} shows a frame in which the stall id is easily visible.   The OCR model we use to identify the numbers in a frame has an accuracy of 99\%. We fine-tune a FasterRCNN model to identify and segment the sub-keyframe. The model achieves an accuracy of 99\% on a given frame. An example of a stall ID is seen in \ref{fig:kf_stallid}. Pseudo-code for the data extraction task appears in Algorithm~\ref{alg:keyAlgo}.

Having selected an image, we organize data about a given cow using a single file system folder per animal, per milking session. To this end, we write a Python program that automatically extracts keyframes, determines the stall ID, creates a suitable folder, and then stores the associated keyframes in that folder. The program obtains frame-by-frame access using the OpenCV package~\footnotemark. To perform teat {\em localization,}, we trained an ML model which entails identifying each of the cow's teats and segmenting them using bounding boxes. Similar to the process used for stall ID identification, line~\ref{alg_is_teat}  uses the function $Loc(teat\_segments)$ to check two things: (1) whether all teats were identified with high confidence; (2) whether the teats are centered in the frame. The first one uses the ML model confidence score to check not only if there are teats in the frame, but also that they were properly segmented. The second one uses the 2D coordinate of the ML result, to ensure that all the teats are captured in the frames, avoiding the frames where just a subset of the  teats were captured. If a frame satisfies both criteria, it is saved to the folder under its stall ID folder, on line~\ref{algo_store_teat}.  

\begin{algorithm}
\caption{Key Frame extraction program}\label{alg:keyAlgo}
\begin{algorithmic}[1]
\Require video\_path, extraction\_rate
\State $cur\_stall\_id \gets null$
\State $folder\_name \gets null$
\State $cap \gets VideoCapture(video\_path)$ \label{alg_vidCapture}
\While{$\exists unprocessed\_frame$}
\State $frame \gets cap.read()$
\State $is\_stall\_key \gets OCR(frame)$ \label{alg_is_stall}
\State $teat\_segments \gets seg\_model(frame)$ 
\State $is\_teat\_key \gets Loc(teat\_segments)$ \label{alg_is_teat}
\If{$is\_stall\_key$}
    \If{$stall\_key \neq cur\_stall\_id$}
        \State $folder\_name = cur\_stall\_id$ \label{alg_change_stall}
        \State $\text{Create\ folder\ with\ folder\_name}$
    \EndIf
\ElsIf{$is\_teat\_key$}
    \State $\text{Store\ teat\_segments\ to\ folder\_name}$ \label{algo_store_teat}
\EndIf
\EndWhile
\end{algorithmic}
\end{algorithm}

\footnotetext{https://opencv.org}

\section{Experimental Evaluation}

\subsection{Model Settings}
For teat health assessment purposes, we consider a set of candidate object detection models.  We select Faster-RCNN \cite{ren2015faster} model as a baseline.  The foundational vision models in this project utilize either convolutional layers or multi-head attention blocks, and sometimes both.  These models are benchmarked in our dataset with different scales to study the trade-off between better model system metrics (run time, memory consumption) and better model performance metrics (validation accuracy and bounding boxes mean average precision for small objects).  We include both two- and single-stage models and will discuss this in the following section.  In the experiment described below, we use mean average precision (mAP) as the performance metric, more specifically, mAP for small objects.  We defer the detailed discussion of the metric in later sections.

\subsection{Fine-tuning the Candidate Models}
Our overall approach is as follows.  First, we undertake an offline process to fine-tune each of the candidate computer vision models using an inexpensive training process that refines the standard model parameters to optimize performance for data collected in our milking parlor.  Next, we expose each tuned model to production data.  The human-expert ground truth labels are used to assess the performance of our automated scoring solutions.

Our work requires models for teat shape identification and teat skin condition classification. We run both tasks on each sub-image (each distinct teat). We consider both two-stage models and single-stage models.  Faster-RCNN \cite{ren2015faster} is a two-stage detector, which relies on a Regional Proposal Networks (RPN) to propose many potential regions of interest (RoI) and then applies a classifier backbone. YOLO-F \cite{chen2021you}, a modified version of YOLO, is a single-stage detector. We then consider the State-Of-The-Art (SOTA) models often observed to have end-to-end transformer architecture.  DINO \cite{zhang2022dino}, a modified version DETR \cite{carion2020end}, uses a transformer architecture. 

Our review of prior research on automated teat condition scoring suggests that the two-stage Faster-RCNN should be viewed as today's best baseline option for teat localization.  We evaluate this baseline both in terms of the scoring performance achieved and the time needed to carry out the scoring procedure:  a rotary milking parlor never stops, and this imposes a form of deadline.

Next, we use our collected and hand-labeled dataset to fine-tune the candidate ML models for cow teat localization and then to optimize skin condition and shape classification within the localization sub-images. We explore ML models under two different network architectures: a two-stage detector and a single-stage detector. Models with two-stage detector architecture, rely on a Regional Proposal Networks (RPN) to propose many potential regions of interest (RoI), and then applies a classifier backbone. Faster-RCNN~\cite{ren2015faster} comes from this setup. Models with single-stage detector architecture merge the two stages into one. Under this architecture, we trained a modified version of YOLO~\cite{redmon2016you}, YOLO-F \cite{chen2021you}.  

Over the past few years, transformers have achieved great success in the vision domain. We select DINO~\cite{zhang2022dino} (a modified version of the first end-to-end object detector, DETR~\cite{carion2020end}) as a candidate transformer-based solution.  

\begin{figure*} [h!]
\centering
\includegraphics[width=17cm]{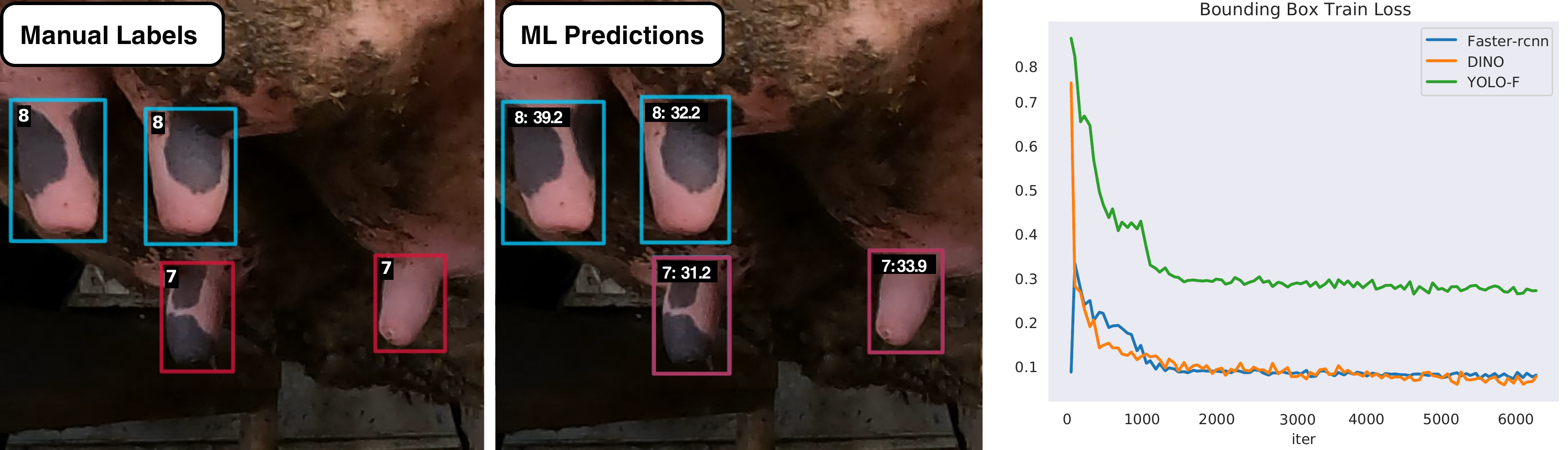} 
\caption{Teat shape images, labels and train loss curve}
\label{fig:Workflow}
\end{figure*}

\subsection{Experimental Results} \label{sec:ExperimentalResults}



All experiments are carried out on an NVIDIA RTX 4090 GPU. We use mAP as our performance metric.  For COCO datasets, mAP is calculated for Intersection over Union(IoU) values.  The IoU is derived by the area of overlap divided by the area of the union in between the ground truth bounding box and the predicted bounding box.  Our dataset consists of only small-scale objects whose areas are often smaller than 32 $\times$ 32 pixels.  So, during training, we focus on the mAPs. For the teat shape identification task, we adopt the aforementioned scoring system and assign one of four class labels $[1, 3, 7, 8 ]$ from worst to best teat shape conditions.  For the skin condition detection, we consider a total of 2 class labels, [C1, C3], with class C3 indicating the existence of skin lesions \cite{evaluationBovine}, and C1 indicating the healthy skin condition. \par

For the model configurations, we use a standard ResNet-50 as the classifier backbone for all three models, while the model scales are rather different.  For Faster-RCNN, if we use a batch of 100 images with an input shape of 2704 $\times$ 1520 $\times$ 3, the model consists of 41.364 million parameters, and it requires 0.208 TFLOPs (tera floating-point operations per second).  For YOLO-F, using the same input shape, the model consists of  42.409 million parameters and requires 98.808 GFLOPs (giga floating point operations per second) to execute.  For DINO, we have a model with 47.546 million parameters and requires 0.274 TFLOPs.


\begin{table} [H]
    \centering
    \begin{tabular}{|c|c|c|}
           \hline
           \textbf{model name} & \textbf{validation mAPs } & \textbf{avg inference time} \\
           \hline
           \textbf{DINO}  & \textbf{0.783} & \textbf{628 ms}\\ 
           \hline
           YOLO-F  & 0.634 & 598 ms\\
           \hline
           Faster RCNN & 0.573 & 576 ms\\
           \hline
    \end{tabular}
    \vspace{-1em}
    \caption{List of Teat Shape model performance, mAPs stands for the bounding boxes mean average precision for small objects}
    \label{tab:result_table}
\end{table}

\vspace{-1em}
\begin{table} [H]
    \centering
    \begin{tabular}{|c|c|c|}
           \hline
           \textbf{model name} & \textbf{validation mAPs } & \textbf{avg inference time} \\
           \hline
           \textbf{DINO}  & \textbf{0.828} & \textbf{505 ms}\\ 
           \hline
           YOLO-F  & 0.615 & 498 ms\\
           \hline
           Faster RCNN & 0.695 & 463 ms \\
           \hline
    \end{tabular}
    \vspace{-1em}
    \caption{List of Teat Skin Condition model performance}
    \label{tab:result_table}
\end{table}

As seen from Table~\ref{tab:result_table}, DINO delivers the best performance and only consumes around 110\% in runtime, compared to the baselines. \par

\begin{figure}[H]
    \centering
    \vspace{-1em}
    \includegraphics[width=9cm]{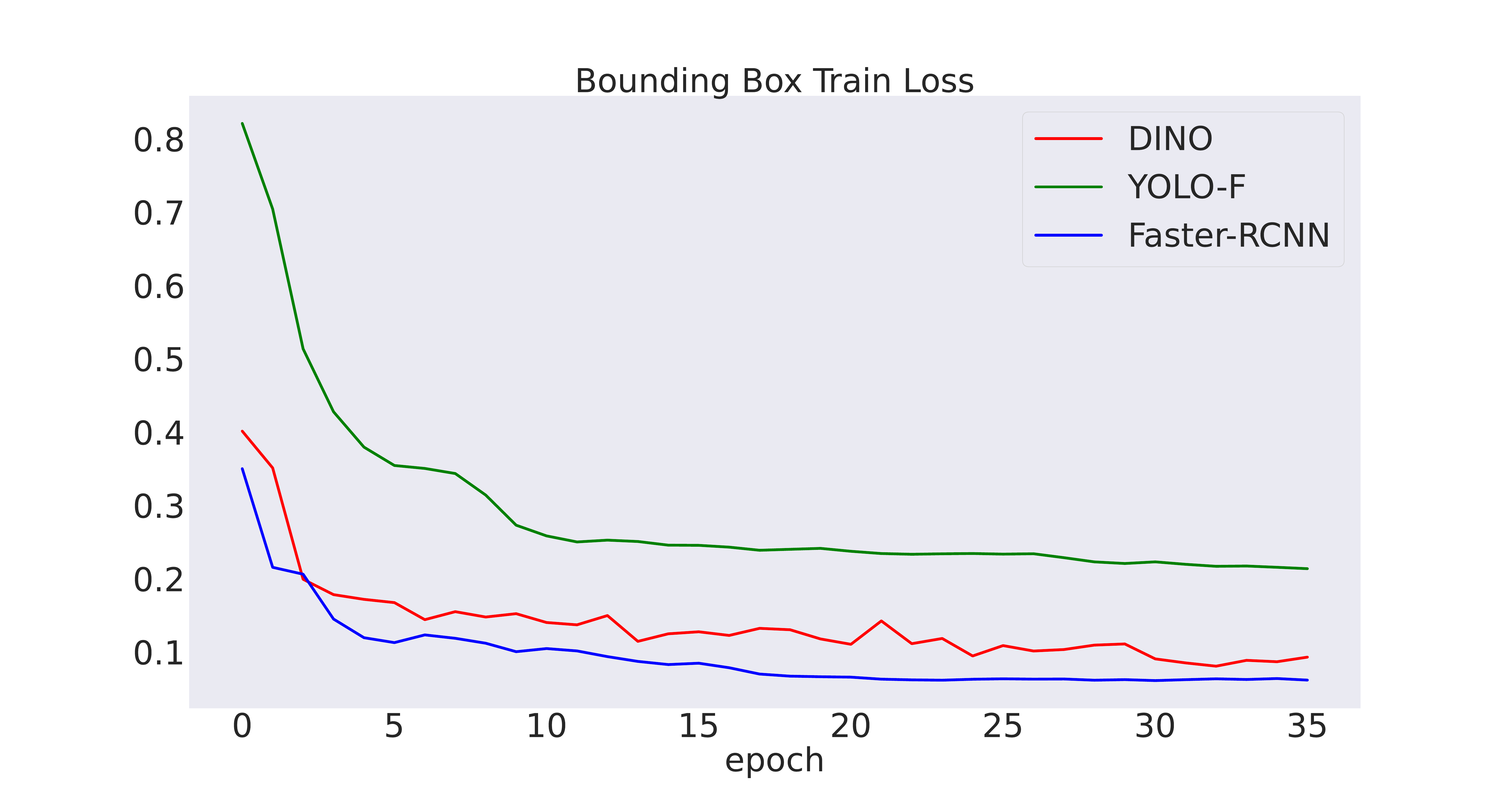}
    \caption{Skin Condition bounding-box training loss curve}
    \label{fig:loss12}
\end{figure}

In figure~\ref{fig:loss12}, we notice that DINO's loss value is actually higher than the loss for our baseline model, and yet DINO outperforms our baseline by around 0.133 when deployed.  Such a finding indicates that the baseline model is prone to overfitting to the training set, becoming a ``narrow specialist'' on training data and yet giving weaker results in actual deployment.    

DINO is slightly slower than other models, but not significantly so.  Indeed, the sub-second performance we obtained is still more than adequate to enable an automated action if a teat health problem is sensed, provided that the computer vision inference task will run physically close to the video camera, with a fast way to access the video data.  Had we deployed our solution on a cloud, delays for uploading video to the cloud could easily have dominated the inference time, but given that our model is small enough to run on a standard laptop, on-premise deployment is reasonable.

\section{Discussion}



\subsection{Efficient Data Storage}
The automatic data processing pipeline described in Section~\ref{sec:key_frame_extraction} transfers the camera-captured video to a keyframe for storage as part of an animal health record and training dataset. One of the reasons for using keyframes instead of raw video is memory efficiency. While raw video contains a lot of information, much of that information is irrelevant to the research, and the video camera continues to run even when there is no cow in the milking parlor.  Moreover, there are circumstances such as the one shown in Figure \ref{fig:wrong_cowteat} 
where the image shows crossed teats, or where one teat obscures another, and hence little can be determined about the condition of the hidden teats.  From a different angle, that same teat might have been clearly visible.

A keyframe is much smaller then a full video clip, and the segmented portion of the frame containing the cow teats even more so. From our collected data, the measured average size per image frame that contains full image with four teats is 800KB on disk. The average size for a segmented teat image is 10KB on the hard disk. In comparison, for a clip of 10-minute raw video that takes 4GB on disk, the distilled keyframe folder is only 139.5MB, whereas  removable intermediate images occupy 581MB. The intermediate images contain the keyframes for stall ID and teat candidate images, from which we choose the one where the teats are centered and clear as the record to store. The memory required to store the raw video file would be almost 28 times more than is required to store the keyframe.

\subsection{Machine Intelligence for Dairy Farms}

ML models can significantly enhance dairy farm health management by operating more efficiently and effectively, capturing nuances that expert veterinarians might miss during long working hours or in an intense farming environment. These roles often involve repetitive teat health scoring tasks. Our duo-camera models can operate 24/7, collecting time series data of teat keyframes. This machine intelligence can provide veterinarians with valuable evidence to support their evaluations and judgments. Furthermore, this technology can be scaled and adapted to other agricultural fields.

We discuss our positive Life Cycle for iteratively improving our model's performance with the improved quality and quantity of data we collected.  We consider a multi-phase setup, where the deliverable for each stage would be deployed to help with further improvement that happens during the next stage.  In particular, we started with a low-data paradigm, where we have quite a limited amount of data, but with high-quality annotation.  We train a model based on this preliminary dataset.  With this model deployed, we were able to automate the process of data collection and remove the unnecessary storage overhead of most video files, and only obtain keyframes.  Our animal scientist would move on to annotate the high-quality raw keyframes.  While we are expanding our dataset, we will be expecting our dataset to incorporate the quantity and quality requirements for developing the ML models.  Additionally, we also argue that with the amount of data we are aggregating, we will be able to automatically eliminate the long-tail distribution of classes that currently exists in our dataset.


\subsection{Limitations \& Future Work}
This paper focuses solely on teat shape identification and skin condition score predictions. However, in future studies, we aim to incorporate additional criteria for teat evaluations, such as predicting teat-end hyperkeratosis scores or assessing udder health in multidimensional teat health analysis. By expanding the scope of teat evaluations, we can achieve a more comprehensive analysis of teat health.\par

Moreover, there is a need for more balanced datasets in AI-based duo-dimensional teat health analysis, particularly due to the scarcity of labels for rare cases. For instance, in our current skin condition dataset, our ratio between normal C1 labels and abnormal C3 labels is 925:44. The unbalanced dataset limits the model to learn from the abnormal situations and impacts model performance. Through large-scale, long-term data collection efforts, we anticipate that our models will demonstrate improved performance in identifying and analyzing these less common labels. In the future work, we plan to collect data from additional farms to ensure more balanced datasets.

We could further investigate additional data augmentation techniques, such as large-scale jittering (LSJ), to enhance image resolutions, camera angles, and lighting conditions, ultimately improving the overall performance of our models. Given that our current datasets were collected under favorable lighting conditions, future large-scale data collection efforts will involve capturing keyframes from diverse environments and implementing methods to enhance image quality.

Our project relies on ground truth labels derived from veterinary expertise. However, teat condition is subjective, hence any single professional could err when scoring, creating a puzzle: if our model is incorrect, did it learn from incorrect training data, or was it confused by poor lighting, animal skin pigmentation, or some other factor?  In situations where ground truth eventually becomes available, techniques such as a confusion matrix (gradient ascent) can offer insights into when and why automation classification errors arise.  This suggests that one could eventually create systems that might dynamically improve their performance, effectively learning from experience. 

\section{Conclusion}
We explore teat localization and shape classification using ML models using a preliminary dataset of 348 images with 968 objects from 4 distinct classes. For teat skin conditions, we generate 946 labels to train ML models for teat health analysis. In this paper, we explore different object detectors across various architectures and found that DINO performs best overall. Our automated digital-twin approach has been shown to yield accurate classifications.  Although our experiments are performed on a size-limited initial dataset, we plan to aggregate a dataset that incorporates both the quantity and quality requirements for developing ML models in the future. 


\appendix

\section{Acknowledgments}
This project is sponsored by the Cornell Institute for Digital Agriculture (CIDA) and receives data support from DeLaval Inc, with sincere appreciation to Mario Lopez and Hayley Hopkins.

\bibliography{aaai24}

\begin{thebibliography}{39}
\providecommand{\natexlab}[1]{#1}

\bibitem[{A~J~Seykora(1985)}]{morphology}
A~J~Seykora, B. T.~M. 1985.
\newblock Udder and teat morphology related to mastitis resistance: a review.
\newblock \emph{Journal of dairy science vol. 68,8}.

\bibitem[{Achour et~al.(2020)Achour, Belkadi, Filali, Laghrouche, and Lahdir}]{visual_identification_feeding_behavior}
Achour, B.; Belkadi, M.; Filali, I.; Laghrouche, M.; and Lahdir, M. 2020.
\newblock Image analysis for individual identification and feeding behaviour monitoring of dairy cows based on Convolutional Neural Networks (CNN).
\newblock \emph{Biosystems Engineering}, 198: 31--49.

\bibitem[{Basran, Wieland, and Porter(2020)}]{BASRAN202010703}
Basran, P.; Wieland, M.; and Porter, I. 2020.
\newblock Technical note: A digital technique and platform for assessing dairy cow teat-end condition.
\newblock \emph{Journal of Dairy Science}, 103(11): 10703--10708.

\bibitem[{Bercovich et~al.(2013)Bercovich, Edan, Alchanatis, Moallem, Parmet, Honig, Maltz, Antler, and Halachmi}]{bcs_dev}
Bercovich, A.; Edan, Y.; Alchanatis, V.; Moallem, U.; Parmet, Y.; Honig, H.; Maltz, E.; Antler, A.; and Halachmi, I. 2013.
\newblock Development of an automatic cow body condition scoring using body shape signature and Fourier descriptors.
\newblock \emph{Journal of Dairy Science}, 96(12): 8047--8059.

\bibitem[{Carion et~al.(2020)Carion, Massa, Synnaeve, Usunier, Kirillov, and Zagoruyko}]{carion2020end}
Carion, N.; Massa, F.; Synnaeve, G.; Usunier, N.; Kirillov, A.; and Zagoruyko, S. 2020.
\newblock End-to-end object detection with transformers.
\newblock In \emph{European conference on computer vision}, 213--229. Springer.

\bibitem[{Chen et~al.(2021)Chen, Wang, Yang, Zhang, Cheng, and Sun}]{chen2021you}
Chen, Q.; Wang, Y.; Yang, T.; Zhang, X.; Cheng, J.; and Sun, J. 2021.
\newblock You only look one-level feature.
\newblock In \emph{Proceedings of the IEEE/CVF conference on computer vision and pattern recognition}, 13039--13048.

\bibitem[{Cockburn(2020)}]{ml_in_dairy_review}
Cockburn, M. 2020.
\newblock Review: Application and Prospective Discussion of Machine Learning for the Management of Dairy Farms.
\newblock \emph{Animals}, 10(9).

\bibitem[{Cook and Reinemann(2007)}]{cook2007tool}
Cook, N.~B.; and Reinemann, D.~J. 2007.
\newblock A tool box for assessing cow, udder and teat hygiene.
\newblock In \emph{annual meeting of the NMC}, 21--24.

\bibitem[{Fadul-Pacheco, Delgado, and Cabrera(2021)}]{mastitis_ml}
Fadul-Pacheco, L.; Delgado, H.; and Cabrera, V.~E. 2021.
\newblock Exploring machine learning algorithms for early prediction of clinical mastitis.
\newblock \emph{International Dairy Journal}, 119: 105051.

\bibitem[{Fauvel et~al.(2019)Fauvel, Masson, Fromont, Faverdin, and Termier}]{ml_estrus}
Fauvel, K.; Masson, V.; Fromont, E.; Faverdin, P.; and Termier, A. 2019.
\newblock Towards Sustainable Dairy Management - A Machine Learning Enhanced Method for Estrus Detection.
\newblock 3051–3059.

\bibitem[{Gorczyca and Gebremedhin(2020)}]{heat_stress_ml}
Gorczyca, M.~T.; and Gebremedhin, K.~G. 2020.
\newblock Ranking of environmental heat stressors for dairy cows using machine learning algorithms.
\newblock \emph{Computers and Electronics in Agriculture}, 168: 105124.

\bibitem[{Gupta et~al.()Gupta, Hao, Yang, Yuan, Wieland, Basran, and Birman}]{gupta_digital_2024}
Gupta, A.; Hao, Y.; Yang, Y.; Yuan, T.; Wieland, M.; Basran, P.~S.; and Birman, K. ????
\newblock Digital Twin-Driven Teat Localization and Shape Identification for Dairy Cow (Student Abstract).
\newblock 38(21): 23510--23511.
\newblock Number: 21.

\bibitem[{Gupta et~al.(2024)Gupta, Hao, Yang, Yuan, Wieland, Basran, and Birman}]{gupta2024digital}
Gupta, A.; Hao, Y.; Yang, Y.; Yuan, T.; Wieland, M.; Basran, P.~S.; and Birman, K. 2024.
\newblock Digital Twin-Driven Teat Localization and Shape Identification for Dairy Cow (Student Abstract).
\newblock In \emph{Proceedings of the AAAI Conference on Artificial Intelligence}, volume~38, 23510--23511.

\bibitem[{Halachmi et~al.(2008)Halachmi, Polak, Roberts, and Klopcic}]{bcs_body_shape}
Halachmi, I.; Polak, P.; Roberts, D.; and Klopcic, M. 2008.
\newblock Cow Body Shape and Automation of Condition Scoring.
\newblock \emph{Journal of Dairy Science}, 91(11): 4444--4451.

\bibitem[{Hillerton(2005)}]{hillerton2005teat}
Hillerton, J.~E. 2005.
\newblock Teat condition scoring-an effective diagnostic tool.
\newblock In \emph{Proc. of National Mastitis Council Regional Meeting}, 37--43.

\bibitem[{Hogeveen et~al.(2011)Hogeveen, Pyorala, Waller, Hogan, Lam, Oliver, Schukken, Barkema, and Hillerton}]{hogeveen2011current}
Hogeveen, H.; Pyorala, S.; Waller, K.~P.; Hogan, J.~S.; Lam, T.~J.; Oliver, S.~P.; Schukken, Y.~H.; Barkema, H.~W.; and Hillerton, J.~E. 2011.
\newblock Current status and future challenges in mastitis research.
\newblock In \emph{Proceedings of the 50th Annual Meeting of the National Mastitis Council, 23-26 January, 2011, Arlington, USA}, 36--48.

\bibitem[{Jiang et~al.(2019)Jiang, Wu, Yin, Wu, Song, and He}]{cow_body_detection}
Jiang, B.; Wu, Q.; Yin, X.; Wu, D.; Song, H.; and He, D. 2019.
\newblock FLYOLOv3 deep learning for key parts of dairy cow body detection.
\newblock \emph{Computers and Electronics in Agriculture}, 166: 104982.

\bibitem[{Joseph~Redmon(2015)}]{redmon2016you}
Joseph~Redmon, R. G. A.~F., Santosh~Divvala. 2015.
\newblock You Only Look Once: Unified, Real-Time Object Detection.
\newblock volume abs/1506.02640.

\bibitem[{Kielland et~al.(2009)Kielland, Ruud, Zanella, and {\O}ster{\aa}s}]{kielland2009prevalence}
Kielland, C.; Ruud, L.; Zanella, A.; and {\O}ster{\aa}s, O. 2009.
\newblock Prevalence and risk factors for skin lesions on legs of dairy cattle housed in freestalls in Norway.
\newblock \emph{Journal of Dairy Science}, 92(11): 5487--5496.

\bibitem[{Lin et~al.(2014)Lin, Maire, Belongie, Hays, Perona, Ramanan, Doll{\'a}r, and Zitnick}]{lin2014microsoft}
Lin, T.-Y.; Maire, M.; Belongie, S.; Hays, J.; Perona, P.; Ramanan, D.; Doll{\'a}r, P.; and Zitnick, C.~L. 2014.
\newblock Microsoft coco: Common objects in context.
\newblock In \emph{Computer Vision--ECCV 2014: 13th European Conference, Zurich, Switzerland, September 6-12, 2014, Proceedings, Part V 13}, 740--755. Springer.

\bibitem[{Lojda and Matouskova.(1976)}]{teatShapOrifice}
Lojda, M.~S., L.; and Matouskova., O. 1976.
\newblock The shape of the teat and teat-end and the location of the teat canal orifice in relation to subclinical mastitis in cattle.
\newblock \emph{Aeta Bet. Brno.}, 181--185.

\bibitem[{M~R, N~K, and V(2022{\natexlab{a}})}]{Disease_Detection}
M~R, S.; N~K, V.; and V, K. 2022{\natexlab{a}}.
\newblock Teat and Udder Disease Detection on Cattle using Machine Learning.
\newblock In \emph{2022 International Conference on Signal and Information Processing (IConSIP)}, 1--5.

\bibitem[{M~R, N~K, and V(2022{\natexlab{b}})}]{teat_udder_ml}
M~R, S.; N~K, V.; and V, K. 2022{\natexlab{b}}.
\newblock Teat and Udder Disease Detection on Cattle using Machine Learning.
\newblock 1--5.

\bibitem[{Mein et~al.(2001)Mein, Neijenhuis, Morgan, Reinemann, Hillerton, Baines, Ohnstad, Rasmussen, Timms, Britt, Farnsworth, and Cook}]{evaluationBovine}
Mein, G.; Neijenhuis, F.; Morgan, W.; Reinemann, D.; Hillerton, E.; Baines, J.; Ohnstad, I.; Rasmussen, M.; Timms, L.; Britt, J.; Farnsworth, R.; and Cook, N. 2001.
\newblock Evaluation of Bovine Teat Condition in Commercial Dairy Herds: 1. Non-Infectious Factors.
\newblock \emph{Proc. Pacific Congress of Milk Quality and Mastitis Control}, 469--478.

\bibitem[{Perez et~al.(2020)Perez, Rubambiza, Barker, Weatherspoon, and Giordano}]{numerical_dairy_system}
Perez, M.; Rubambiza, G.; Barker, B.; Weatherspoon, H.; and Giordano, J. 2020.
\newblock Automated real-time integration of data from multiple sensors and nonsensor systems for prediction of dairy cow and herd status and performance.
\newblock 103: 119--119.

\bibitem[{Porter, Wieland, and Basran()}]{porter_feasibility_2021}
Porter, I.~R.; Wieland, M.; and Basran, P.~S. ????
\newblock Feasibility of the use of deep learning classification of teat-end condition in Holstein cattle.
\newblock 104(4): 4529--4536.

\bibitem[{Ren et~al.(2015)Ren, He, Girshick, and Sun}]{ren2015faster}
Ren, S.; He, K.; Girshick, R.; and Sun, J. 2015.
\newblock Faster r-cnn: Towards real-time object detection with region proposal networks.
\newblock \emph{Advances in neural information processing systems}, 28.

\bibitem[{Ruegg(2003)}]{ruegg2003investigation}
Ruegg, P.~L. 2003.
\newblock Investigation of mastitis problems on farms.
\newblock \emph{Veterinary Clinics: Food Animal Practice}, 19(1): 47--73.

\bibitem[{Russell et~al.(2008)Russell, Torralba, Murphy, and Freeman}]{russell2008labelme}
Russell, B.~C.; Torralba, A.; Murphy, K.~P.; and Freeman, W.~T. 2008.
\newblock LabelMe: a database and web-based tool for image annotation.
\newblock \emph{International journal of computer vision}, 77: 157--173.

\bibitem[{Rutten et~al.(2013)Rutten, Velthuis, Steeneveld, and Hogeveen}]{behavioral_ml}
Rutten, C.; Velthuis, A.; Steeneveld, W.; and Hogeveen, H. 2013.
\newblock Invited review: Sensors to support health management on dairy farms.
\newblock \emph{Journal of Dairy Science}, 96(4): 1928--1952.

\bibitem[{Schreiner and Ruegg(2003)}]{schreiner2003relationship}
Schreiner, D.; and Ruegg, P. 2003.
\newblock Relationship between udder and leg hygiene scores and subclinical mastitis.
\newblock \emph{Journal of dairy science}, 86(11): 3460--3465.

\bibitem[{Seykora and McDaniel(1985{\natexlab{a}})}]{seykora_heritabilities_1985}
Seykora, A.; and McDaniel, B. 1985{\natexlab{a}}.
\newblock Heritabilities of Teat Traits and their Relationships with Milk Yield, Somatic Cell Count, and Percent Two-Minute Milk1.
\newblock \emph{Journal of Dairy Science}, 68(10): 2670--2683.

\bibitem[{Seykora and McDaniel(1985{\natexlab{b}})}]{seykora1985udder}
Seykora, A.; and McDaniel, B. 1985{\natexlab{b}}.
\newblock Udder and teat morphology related to mastitis resistance: a review.
\newblock \emph{Journal of Dairy Science}, 68(8): 2087--2093.

\bibitem[{Slob, Catal, and Kassahun(2021{\natexlab{a}})}]{ml_in_dairy}
Slob, N.; Catal, C.; and Kassahun, A. 2021{\natexlab{a}}.
\newblock Application of machine learning to improve dairy farm management: A systematic literature review.
\newblock \emph{Preventive Veterinary Medicine}, 187: 105237.

\bibitem[{Slob, Catal, and Kassahun(2021{\natexlab{b}})}]{slob2021application}
Slob, N.; Catal, C.; and Kassahun, A. 2021{\natexlab{b}}.
\newblock Application of machine learning to improve dairy farm management: A systematic literature review.
\newblock \emph{Preventive Veterinary Medicine}, 187: 105237.

\bibitem[{Spoliansky et~al.(2016)Spoliansky, Edan, Parmet, and Halachmi}]{bcs_3d}
Spoliansky, R.; Edan, Y.; Parmet, Y.; and Halachmi, I. 2016.
\newblock Development of automatic body condition scoring using a low-cost 3-dimensional Kinect camera.
\newblock \emph{Journal of Dairy Science}, 99(9): 7714--7725.

\bibitem[{Wieland, Nydam, and Virkler(2017)}]{wieland2017longitudinal}
Wieland, M.; Nydam, D.; and Virkler, P. 2017.
\newblock A longitudinal field study investigating the association between teat-end shape and two minute milk yield, milking unit-on time, and time in low flow rate.
\newblock \emph{Livestock Science}, 205: 88--97.

\bibitem[{Wieland et~al.(2018)Wieland, Nydam, Älveby, Wood, and Virkler}]{WIELAND201811447}
Wieland, M.; Nydam, D.; Älveby, N.; Wood, P.; and Virkler, P. 2018.
\newblock Short communication: Teat-end shape and udder-level milking characteristics and their associations with machine milking-induced changes in teat tissue condition.
\newblock \emph{Journal of Dairy Science}, 101(12): 11447--11454.

\bibitem[{Zhang et~al.(2022)Zhang, Li, Liu, Zhang, Su, Zhu, Ni, and Shum}]{zhang2022dino}
Zhang, H.; Li, F.; Liu, S.; Zhang, L.; Su, H.; Zhu, J.; Ni, L.~M.; and Shum, H.-Y. 2022.
\newblock Dino: Detr with improved denoising anchor boxes for end-to-end object detection.
\newblock \emph{arXiv preprint arXiv:2203.03605}.

\end{thebibliography}

\end{document}